\documentclass[a4paper,twoside]{article}
\pdfoutput=1

\usepackage{epsfig}
\usepackage{subcaption}
\usepackage{calc}
\usepackage{amssymb}
\usepackage{amstext}
\usepackage{amsmath}
\usepackage{amsthm}
\usepackage{multicol}
\usepackage{pslatex}
\usepackage{apalike}
\let\bibhang\relax
\usepackage{algorithm2e}
\usepackage[bottom]{footmisc}
\usepackage{url}
\usepackage{adjustbox}
\usepackage{hyperref}
\usepackage{natbib}
\usepackage[capitalize,noabbrev]{cleveref}
\usepackage{SCITEPRESS}     
\usepackage[T1]{fontenc}

\newcommand{\ballervec}{\texttt{baller2vec}}
\newcommand{\engine}{\textit{SportsNGEN}}

\usepackage{enumitem}
  \newlist{inlinelist}{enumerate*}{1}
  \setlist*[inlinelist,1]{%
          label=(\roman*),
      }

\begin{document}

\title{SportsNGEN: Sustained Generation of Realistic Multi-player Sports Gameplay}


\author{\authorname{Lachlan Thorpe\sup{1}, Lewis Bawden\sup{1}, Karanjot Vendal\sup{1}, John Bronskill\sup{2} and Richard E Turner\sup{2}}
\affiliation{\sup{1}Hawk-Eye Innovations Ltd. \sup{2}University of Cambridge}
\email{\{lachlan.thorpe, lewis.bawden, karanjot.vendal\}@hawkeyeinnovations.com, \{jfb54, ret26\}@cam.ac.uk}
}



\abstract{
We present a transformer decoder based sports simulation engine, \engine{}, trained on sports player and ball tracking sequences, that is capable of generating sustained gameplay and accurately mimicking the decision making of real players.
By training on a large database of professional tennis tracking data, we demonstrate that simulations produced by \engine{} can be used to predict the outcomes of rallies, determine the best shot choices at any point, and evaluate counterfactual or \textit{what if} scenarios to inform coaching decisions and elevate broadcast coverage.
By combining the generated simulations with a shot classifier and logic to start and end rallies, the system is capable of simulating an entire tennis match.
We evaluate \engine{} by comparing statistics of the simulations with those of real matches between the same players.
We show that the model output sampling parameters are crucial to simulation realism and that \engine{} is probabilistically well-calibrated to real data.
In addition, a generic version of \engine{} can be customized to a specific player by fine-tuning on the subset of match data that includes that player.
Finally, we show qualitative results indicating the same approach works for football.
}

\onecolumn \maketitle \normalsize \setcounter{footnote}{0} \vfill
\begin{figure*}[t]
\centering
\includegraphics[width=0.80\textwidth]{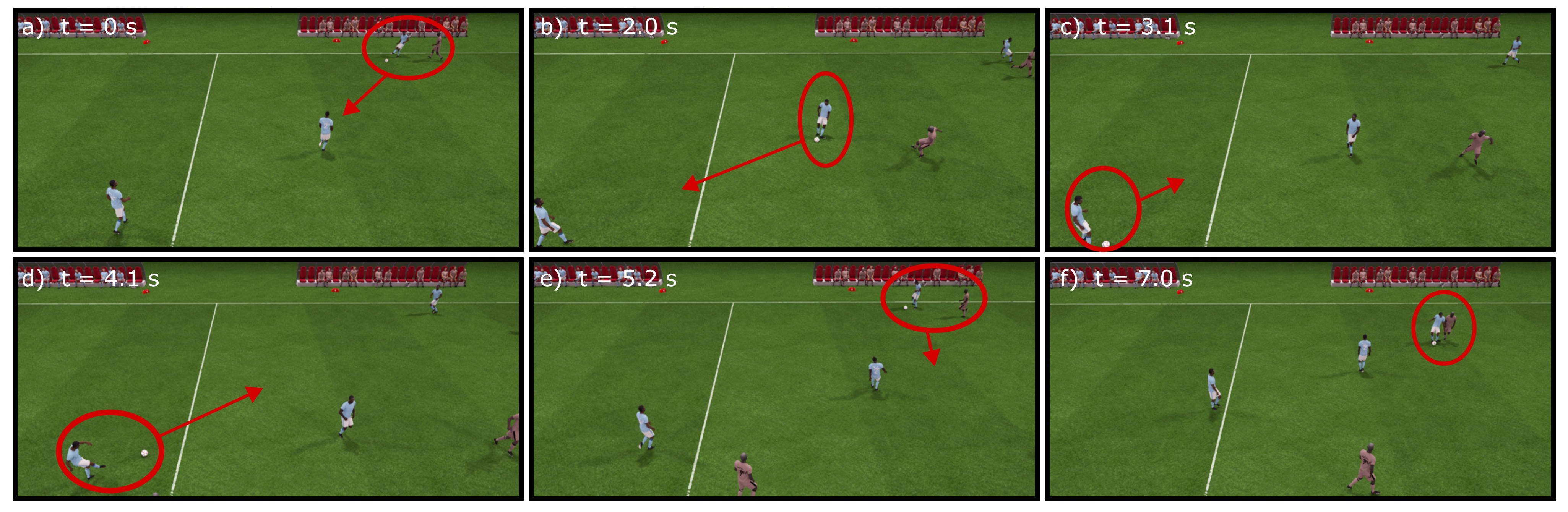}
\caption{Frames from a football match simulated using \engine{}. The panels depict a passing sequence involving 3 players. The ball is in the red circle, with an arrow depicting the play that follows. Link to video: \url{https://youtu.be/M0kkKiGVNzk}}
\label{fig:filmstrip}
\end{figure*}

\begin{figure*}[h]
\centering
\includegraphics[width=0.80\textwidth]{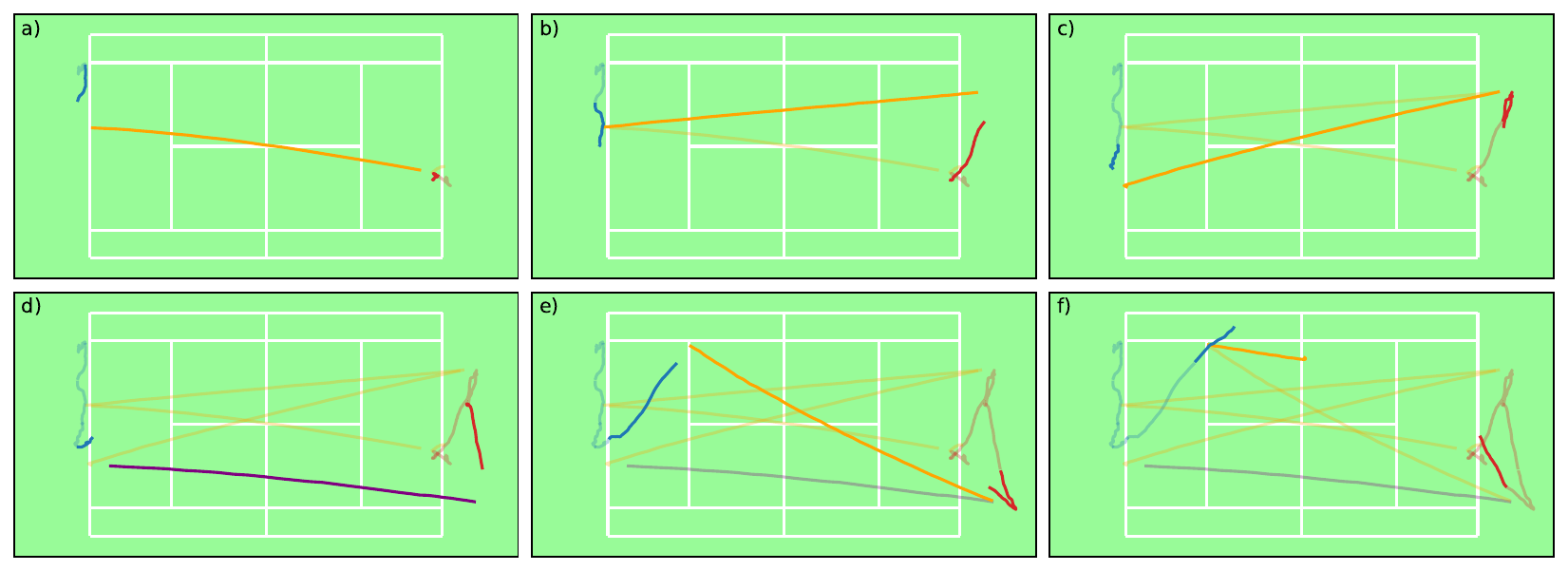}
\caption{Simulated tennis rally between 2 players using 3 shots of training data as input. Frames a) - c): Training data shots. Frames d) - f) Simulated rollout. Red and blue markings indicate player movement. The lines indicate shot trajectories. The current shot is opaque while earlier shots are more transparent. The purple line is the first simulated shot. Link to video: \url{https://youtu.be/A1_vv12V5q0}}
\label{fig:rollouts_shot_decision_multipanel}
\end{figure*}
\section{Introduction}
\label{Introduction}
The application of machine learning methods has proven beneficial to many sports applications \citep{zhao2023survey}. In particular, sports simulation and analysis can provide valuable insights to sports teams when attempting to understand how small changes to player formation or playing style could impact the next period of play, or their chances of winning \citep{hauri2022group, teranishi2022evaluation, wang2023tacticai}. In addition, realistic gameplay simulation is critical in computer gaming scenarios \citep{kurach2020google}.

Tremendous progress has been made in the area of sports trajectory prediction \citep{yue2014learning, zheng2016generating, le2017coordinated, zhan2019generating, li2021grin, tang2021collaborative, wu2021basketball, alcorn2021baller2vec, omidshafiei2022multiagent}, however it is difficult to precisely mimic training data over long periods of time.
\cref{fig:mse_vs_time} shows how the prediction error of the player and ball positions increases with time when simulated tennis data from our system is compared to the training data.
Sports are inherently unpredictable over time scales longer than several seconds and as a result, deterministic prediction is not possible or useful in many scenarios.
Instead, it is important to capture the different ways a match will evolve in a statistically accurate manner by modelling the complete distribution of player decision making.

We propose that generated sports simulations should be:
\begin{inlinelist}
\setlength\itemsep{0pt}
    \item highly \textit{realistic} and capture the complete distribution of real player behaviour;
    \item \textit{sustained} for the duration between natural breaks in the gameplay;
    \item \textit{customizable} via fine-tuning or other method to emulate the style of play of a particular player and/or team; and
    \item \textit{measurable} in that metrics are available to evaluate the quality of the simulations (as opposed relying on a human expert) such that the simulations can be improved by optimizing the metrics.
\end{inlinelist}

Recently, the transformer architecture \citep{vaswani2017attention} has been applied to multi-agent spatiotemporal systems problems to generate realistic sports simulations and understand player behavioural patterns \citep{alcorn2021baller2vec}.
Instead of generating words as in natural language processing, temporal player and ball movements can be generated by training a transformer model to predict the next position from a sequence of tracking data.

However, to the best of our knowledge, no previous work has been successful in generating realistic, sustained, and customizable simulations, learned from player and ball tracking data, for more than a few seconds.
In this work we present Sports Neural Generator or \engine{} that realizes the goals of realistic, sustained, customizable and measurable sports gameplay.
\cref{fig:filmstrip} and \cref{fig:rollouts_shot_decision_multipanel} depict football\footnote{We use the term football to refer European football or soccer.} and tennis sequences, respectively, generated by our approach along with links to simulation videos.

Our contributions:
\begin{inlinelist}
\setlength\itemsep{0pt}
    \item A transformer decoder based simulation engine, \engine{}, trained on player and ball tracking data as well as match metadata, capable of simulating the movement of all players and the ball simultaneously in a sports game scenario. The simulations are sustained between breaks in play.
    \item Training and evaluating \engine{} on a large database of professional tennis tracking data. \engine{} is capable of simulating an entire tennis match by combining the generated simulations with a shot classifier and logic to start and end rallies. 
    \item We show that our model can be used to inform tennis coaching decisions and best shot options by evaluating counterfactual or \textit{what if} options.
    \item We demonstrate through ablations that the following enhancements significantly improve convergence and generated simulations: a) extending the player and ball representations to include relative velocity, distance to the ball, and time into the game or sequence; b) adding small perturbations to the ball positions during training to allow the model to correct for errors; and c) Adding context tokens to allow the model to adapt to different playing surfaces.
    \item We devise a novel optimization method by defining metrics to statistically evaluate the quality of generated tennis data. By altering simulation hyperparameters, we show that the simulations can be optimized to be statistically similar to the behaviour of real players.
    \item We demonstrate that a generic version of our model can be customized to a specific tennis player by fine-tuning on match data that includes that player.
    
\end{inlinelist}
\section{Related Work}
\label{sec:related_work}
In this section, we discuss related work in the categories of sports analytics, and game simulation, and trajectory prediction.

\paragraph{Group Activity Recognition and Sports Analytics}
\citet{pmlr-v32-miller14} develop an approach to represent and analyze the underlying spatial structure that governs shot selection among professional basketball players.
\citet{le2017data} employ an imitation learning approach to analyze football defensive strategies.
\citet{hauri2022group} propose a transformer-based architecture with a  Long Short-Term Memory (LSTM) embedding to recognize basketball group activities from player and ball tracking data.
\citet{teranishi2022evaluation} evaluate football players who create off-ball scoring opportunities by comparing actual movements with the reference movements generated via trajectory prediction.
\citet{chen2023professional} use a probabilistic diffusion approach to model basketball player behavior.
The model only considers player movement and no other metadata.
\citet{wang2023tacticai} present a football tactics assistant that focuses on analyzing corner kicks which allows coaches to explore player setup options and use those with the highest likelihood of success.

\paragraph{Game Simulation}
\citet{kurach2020google} introduce a game engine that simulates football gameplay with an environment for evaluating RL algorithms.
\citet{liu2021motor} demonstrate an RL approach, where the agents progressively learn to play football initially from random behavior, to simple ball chasing, to showing evidence of cooperation.
\citet{braga2022rsoccer} introduce a simulator for robot football optimized for performing RL experiments.
\citet{yu2023asynchronous} introduce a RL environment where agents are trained to play basketball.

Finally, \citet{yuan2023learning} describe a method to learn simulated tennis skills from broadcast videos. However, this approach only models one shot cycle at a time, using statistical analysis to predict the desired shot location for specific players.
There is no coupling between the current shot and previous shots, so strategic play is limited.
Also, the players are only allowed to move on the baseline, so no volleys or inner court play is permitted, restricting realism.

Our approach is not RL based.
It instead learns in a discriminative fashion from sequences of gameplay tracking data, which obviates the need to use physics based models of gameplay or learning gameplay from scratch with RL.
This also enables us to build predictive models for specific players which can be important for analysis and gaming scenarios.
Overall, our work is distinct from the above works in that our goal is to generate sports gameplay that captures the complete distribution of player behaviour, where the aggregation of non-deterministic simulations is statistically similar to real data. This acts as a powerful tool for strategic analysis, evaluating how player decision making affects the outcome of a period of play.

\paragraph{Sports Trajectory Prediction}
There is a rich literature on trajectory prediction in general, and sports trajectory prediction in particular.
\citet{yue2014learning} learn predictive models for basketball play prediction given the current game state.
\citet{zheng2016generating} model spatiotemporal trajectories over long time horizons using expert demonstrations capable of generating realistic, but short rollouts.
\citet{le2017coordinated} present an LSTM based imitation learning approach for learning multiple policies for team defense in professional football. However, no policy is learned for the position of the ball.
\citet{zhan2019generating} describe a hierarchical framework for sequential generative modeling that can generate high quality trajectories and encode coordination between agents.
However, their framework cannot generate entire games.
\citet{li2021grin} describes an approach for multi-agent trajectory prediction using a graph neural network.
When evaluated on basketball data, only short trajectories were considered.
\citet{tang2021collaborative} propose the concept of collaborative uncertainty, to model the uncertainty in interaction in multi-agent trajectory forecasting.
\citet{wu2021basketball} propose a generative adversarial network (GAN) to generate short basketball player and ball trajectories.
\citet{alcorn2021baller2vec} introduce \ballervec{}, a multi-entity transformer that can model coordinated agents.
It employs a special self-attention mask to learn the distributions of statistically dependent agent trajectories and is shown to generate realistic trajectories for basketball players or the ball itself for short durations.
Our work builds upon \ballervec{} to enable sustained gameplay simulations by simultaneously simulating both the player and the ball.
\citet{omidshafiei2022multiagent} study the problem of multiagent time-series imputation in the context of football in order to predict the behaviors of off-screen players.
\section{Methodology}
\label{sec:methodology}
\begin{figure}[t]
\begin{center}
\centerline{\includegraphics[width=0.9\columnwidth]{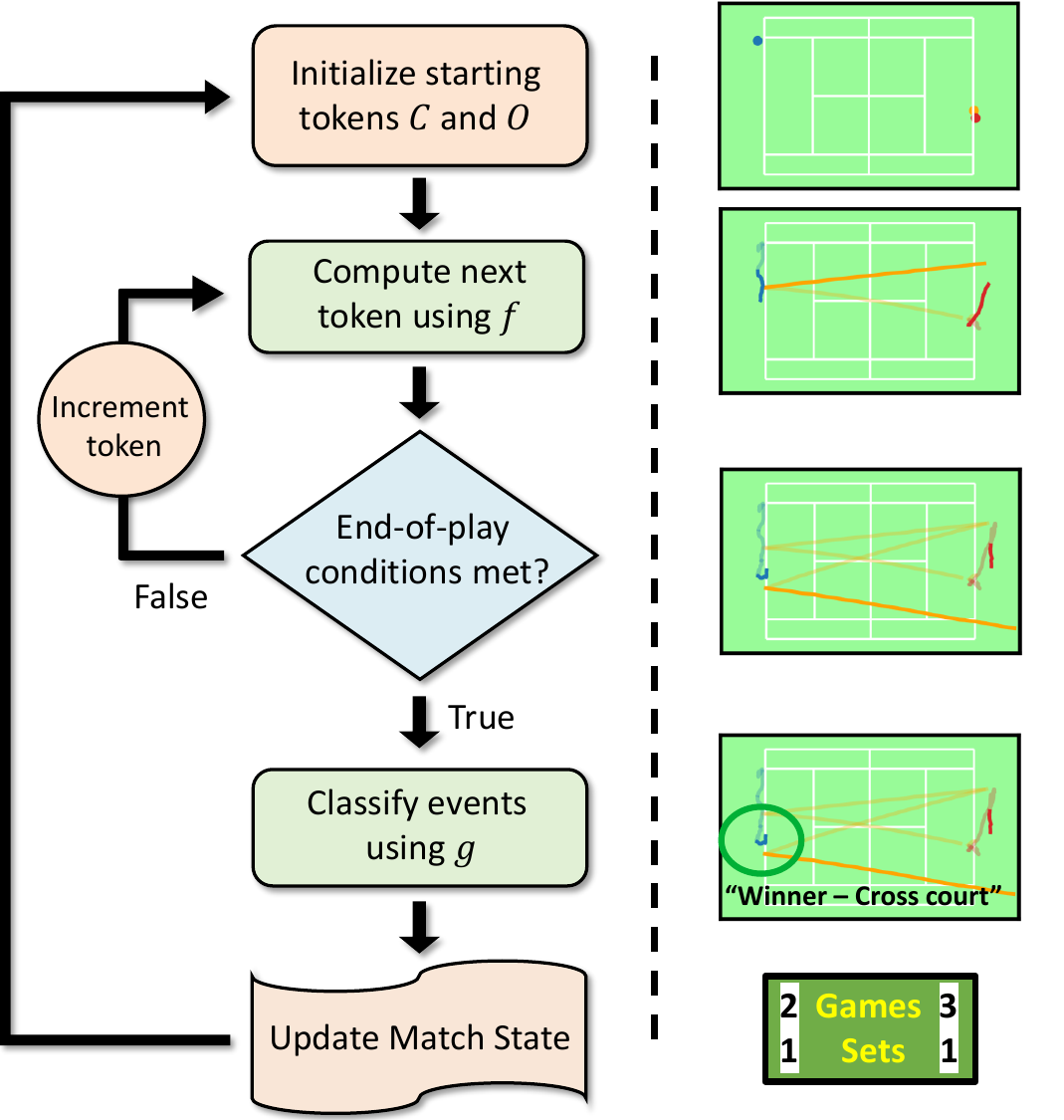}}
\caption{Left: \engine{} flow diagram. Right: Cartoons from a simulated tennis match corresponding to the flow chart steps.}
\label{fig:match_engine}
\end{center}
\end{figure}

In this section we provide a complete description of our approach to generating sports simulations.
A flow diagram of \engine{} is shown in \cref{fig:match_engine}.
\subsection{Input Data}
\label{sec:input_data}
\begin{figure}[ht]
\begin{center}
\centerline{\includegraphics[width=\columnwidth]{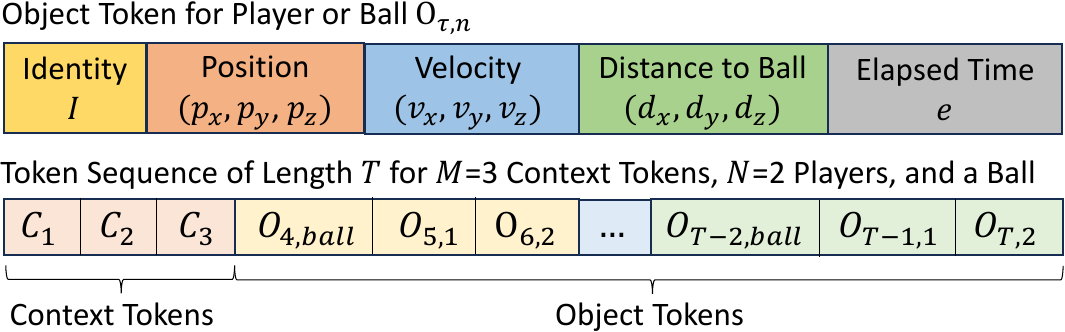}}
\caption{Top: Layout of an object token $O_{\tau, n}$. Bottom: Sequence of $T$ tokens for $M$=3 context tokens, $N$=2 players, and a ball.}
\label{fig:input_description}
\end{center}
\end{figure}
We index the $N$ players and the ball in a match with $n \in \{1, \ldots, N, \texttt{ball}\}$.
We then define an \textit{object token} $O_{\tau, n}$ at index $\tau$ to represent the state of $n$th player or ball as:
\begin{align*}
 O_{\tau,n} = \{I_n, (p_{x,\tau,n}, p_{y,\tau,n}, p_{z,\tau,n}), (v_{x,\tau,n}, v_{y,\tau,n}, v_{z,\tau,n}),\\ (d_{x,\tau,n}, d_{y,\tau,n}, d_{z,\tau,n}), e \}  
\end{align*}
where $p$ denotes position, $v$  velocity, $d$  distance to the ball, $I \in \mathbb{R}^\iota$ a learned identity for a player that can capture their style of play, $e \in \mathbb{R}$ elapsed time into the game or sequence depending on the sport, and $x,y,z \in \mathbb{R}^3$ are components in a 3D coordinate system.
The position data are typically supplied as the center of mass (COM) of the ball or player from a sports tracking system.
For all players, position is 2D only i.e. $p_{z,\tau,n} = v_{z,\tau,n} = 0$ and for the ball, distance $d$ is set to 0. 
The $e$ component of the feature vector is useful to model long-term dependencies due to player fatigue and team strategy or for ensuring simulated tennis rallies are realistic in length.
We normalize the $p$, $v$, and $d$ components of $O$ by appropriate values for each sport.

As a crucial step in generating sustained simulations, we add a small amount of uniform noise to the position $p$ and velocity $v$ of the ball.
We find that training on noise-free ball trajectories does not lead to stable simulations as any errors in the prediction lead to out-of-distribution inputs at the next time step, which the model cannot correct.

In addition to the object tokens, we also define a set of \textit{context tokens} $\{C_1, \ldots, C_M\}$ specific to each sport that contain information that would influence gameplay such as the score, the identity of the opposing team, the location of the game, and the weather.
We convert each piece of contextual information into feature vectors, either through learned encodings for discrete information such as the stadium, or training a network to convert a representation of the score into a feature vector.
\cref{fig:input_description} depicts the components of a token and the order of tokens in a training sequence.
\paragraph{Cropping Sequences}
\label{Methodology - Cropping Sequences}
We crop the input training sequences to eliminate data outside of actual gameplay. The data removed includes players getting into position for the next play or switching sides which are not essential for simulation.
To train the model efficiently using batches, we define a maximum sequence length of tokens $T$ and cut any sequences longer than this into multiple sequences.
Shorter sequences are padded to make up the remainder of the maximum length.
The sequence length $T$ depends on the sample rate of the data, and the length of previous data relevant to predicting the next time step.
Tracking data can be sampled up to 50 Hz.
Although this provides extremely fine detail, for team sports like football with 23 objects on the pitch, a period of 5 seconds at 50 Hz would produce a sequence length of 5750 tokens, making the model impractical to train.
Since many of the dynamics in matches are longer than 5 seconds, we make a compromise between sample rate and computational cost.
\subsection{Transformer Decoder Model}
We use a transformer decoder model $f$ that is an extended implementation of \ballervec{} \citep{alcorn2021baller2vec} to predict future player and ball states given the current and recent history of states. We make a significant update to the \ballervec{} experiments by modelling both the ball and the players simultaneously. 

The model $f$ is run in an auto-regressive mode with a rolling window of length $T$, using a specified period of previous predictions to predict the ball and player state at the next step.
We use the same attention method as \ballervec{}, permitting each object token to attend to every object token up to and including its own time step.
We adjust the attention mask so that each object token can attend to the context tokens, influencing the predictions for player and ball movement.

\begin{figure}[ht]
\begin{center}
\centerline{\includegraphics[width=0.45\textwidth]{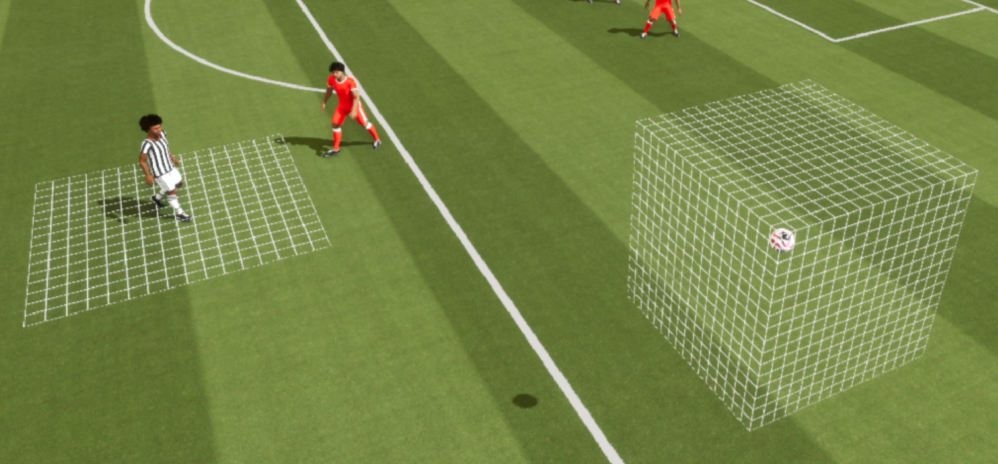}}
\caption{Visualization of the 2D and 3D classification grids used to predict the position of a player and the ball at the next time step.}
\label{fig:classification_grid}
\end{center}
\end{figure}

We treat the update step as a classification as opposed to a regression or diffusion problem, by splitting the area of possible next locations for the ball and players into a 3D and 2D grid, respectively, of discrete bins that indicate the relative offset $\rho$ from the current position $p$ as this is easier to learn and can bound motion to physically possible values.
A depiction of a grid for a football player and the ball is shown in \cref{fig:classification_grid}.

We use nucleus sampling \citep{holtzman2020curious} to sample the location in the output grid based on the output probabilities of $f$.
When the grid location has been selected, we turn the discrete value into a continuous value by sampling from a uniform distribution across the bin.
If the initial conditions for the player or ball have zero velocity, this helps to force the simulation into motion by avoiding continuous velocity predictions of zero.

To enable the model to learn the behavior of individual players, the bin size must be fine grained enough for predictions to capture distinguishing features.
In many sports, important statistics include how fast a player can run, or how far they can hit, throw or kick the ball.
Formally, the probability distribution of predicting a particular bin location $k$ for an object $n$ at step $\tau + 1$ is
\begin{align*}
    p(\rho_{\tau + 1, n} = k | O_{1:\tau,n}) = f(O_{1:\tau,n},k).
\end{align*}
The value of $\rho$ is then sampled from the distribution:
\begin{align*}
    \rho_{\tau + 1,n} \sim p(\rho_{\tau + 1,n} = k | O_{1:\tau,n}).
\end{align*}
Based on the sampled value of $\rho$ and the mapping between bins and physical distance, the updated values of position $p_{\tau + 1,n}$, velocity $v_{\tau + 1,n}$, and distance to the ball $d_{\tau + 1,n}$ can be computed.
Since we use the \ballervec{} attention mask, the positions of the ball and each player can be updated simultaneously at each time step.

We detect the end of a simulation or break in a play with logic specific to each sport.
For example, we can end simulations if a ball goes out of bounds or in some sports if the ball makes contact with the ground, or if the time in the period of play runs out.
When generating simulations, we set a maximum input sequence length of $T$ tokens.
 For a player and ball state update at step $\tau +1$, we input from $\tau - T$ to $\tau$ steps of initial token data into the model $f$. If $T$ time steps of data are not yet simulated, the missing tokens are padded with zeros and masked. Specifically, simulations are rolled autoregressively out at the $i$th step as
 \begin{align*}
     \rho_{i} \sim p(\rho_{i} = k | O_{i-1}, O_{i-2} \ldots O_{i - T}).
 \end{align*}
\subsection{Event Classification}
We also train an event classifier $g$ which is run after a break in gameplay.
Examples of events would be passes, runs, fouls, goals, the type of shot played, and so on.
The event classifier $g$ has the same input and architecture as $f$, but does not use attention masking, and uses separate prediction heads for each different type of event.
The event classifier can be used for defining the initial conditions for the next play and gathering statistics about the period of play.

\section{Tennis Implementation Details}
\label{sec:tennis_implementation}
In this section, we detail the implementation of \engine{} for tennis.
Initial rally conditions, boundary logic and relevant player statistics are well defined, so we can demonstrate the capabilities of the system.

We use a proprietary dataset of tennis tracking data.
The data contains COM locations for each player and the ball sampled at 25 Hz, with the center of the court at $(x,y,z)=(0,0,0)$, whose components refer to the length, width, and vertical directions, respectively.

The data also contains metadata about each match and rally, including: the players in the rally, the tournament and court, the rally winner, whether the rally was a first or second serve, and what shots were played. 
The tracking data set is cut up into individual sequences that start at the toss before a serve and end shortly after the rally is finished. We set a maximum sequence length of input data to be 6 seconds.
We found that increasing the sequence length to be more than 6 seconds became computational impractical and did not improve the model accuracy. This suggests that professional tennis players' decision making is not strongly affected by information further than 6 seconds into the past.
We also double the size of the data set by flipping the data along the $x$ and $y$ axes simultaneously.

We allow for $\pm$25 mm of uniform position and per unit time velocity noise in the $x$ dimension and $\pm$12.5 mm of noise in the $y$ and $z$ dimensions.
If the added noise is any smaller than this, the simulations start to break down.
For output classification, we use 61 bins for each dimension, scaled for the ball such that the maximum velocity is fractionally faster than the current fastest serve speed.
This results in 61 $\times$ 61 = 3721 and 61 $\times$ 61 $\times$ 61 = 226981 possible bin locations for the player and ball output, respectively.
At $25$ Hz, this equates to a ball bin size of $\{x,y,z\}=\{46, 13, 10\}$ mm.

The playing surface is important contextual information when predicting rallies in tennis.
The expectation is that hard and grass courts have the fastest bounces, and clay courts absorb more momentum from the impact resulting in slightly slower bounces and longer rallies.
We learn context vectors for each surface and tournament in the dataset, and also encourage the model to learn the difference between first and second serve types by including context vectors for both.

We generate initial conditions based on historical examples from the data when particular players are serving first or second serves from specific sides.
We take the initial condition as the start of the toss movement during the serve.
This initial condition includes the positions and velocities in all dimensions for both players and the ball.
We can detect the end of the rally through simple logic on the movement of the ball.
If the ball continues past a player, is close to stationary near the net, bounces out of bounds or bounces twice on one side of the net, then we can deem the rally to have finished. 
At this point, we stop the simulation and collect the rally data using the event classifier. 

To understand who won the rally, and for analysis of the point, we train the event classifier to classify the type of shot being played at every step within the simulation.
This includes the type of stroke (groundstroke, serve, volley, etc.), the direction of the shot (cross court, down the line, etc.), whether the shot is a winner, error, or a continuation of the point, and if an error is forced or unforced.

The event classifier $g$ receives as input a simplified version of the input token, without any identity $I$ or context $C$ components.
In the training data, shot type labels are consistent across time steps between shots.
The model is expected to predict the same, only varying its prediction when the ball contacts a racket.
When a rally is finished, we convert the tracking data from the rally into the shot type classifier input, run the model once to identify where the changes in shot type are, and take the model shot type between changes as the final label for each shot.
The winner or error classification for the final shot of the rally tells us who won the point, and the shot type labels help us break down the shots for statistical analysis.
To combine rallies together to simulate an entire match, all that is left to do is implement logic to increment the score, calculate who is serving, from which end and which side.
These can be used to obtain the initial conditions for the next point.

\paragraph{Tennis Network Architecture}
The input tokens $O_{\tau, n}$ are embedded with a 3 layer MLP with input size 30, hidden sizes 256 and 512, and output size 2048.
The transformer decoder, $f$,  has 4 layers, 2048 embedding dimension, 8 heads, 4 expansion factor, and 0.2 dropout.
The shared player output network is a single linear layer with input size 2048 output size equal to 61 $\times$ 61 bins.
The ball output network is a single linear layer with input size 2048 and output size equal to 61 $\times$ 61 $\times$ 61 bins.
%
%
%
\section{Experiments}
\label{sec:experiments}
For the tennis experiments, we selected 3 male professional players with varying styles to evaluate \engine{} and simulated 6 matches between each combination of two players. Each match was the best of 3 sets.
We repeat this experiment across 3 different tournaments, one for each surface type: hard, clay and grass.
For comparison, we then collect data where these players have played each other on these surfaces.
Using both real and simulated data, we compute relevant statistics and define an evaluation metric for each statistic as the difference between the two.

For physical metrics, we compare the median, inter-quartile range (IQR), and Wasserstein distance between the distributions of real and simulated data for the following quantities collected across all matches: 
\begin{itemize}
    \item \textbf{Toss contact height}: Height of the ball at the contact point with the racket during serving.
    \item \textbf{First and second serve speeds}: Maximum recorded speed during the serve.
    \item \textbf{Return speeds}: Maximum speed of a return of serve.
    \item \textbf{Groundstroke speeds}: Maximum speed of all groundstrokes.
\end{itemize}

We also compute additional relevant statistics based on aggregated data. For these quantities, a scalar value is aggregated over many rallies for each player. The absolute difference between the real and simulated aggregated scalars is compared.
\begin{itemize}
    \item \textbf{First serve $\%$}: Percentage of first serves that are in bounds.
    \item \textbf{Double fault $\%$}: Percentage of second serves that are out of bounds.
    \item \textbf{First and second serve win $\%$}: Percentage of rallies won when serving on first and second serve, respectively.
    \item \textbf{Ace $\%$}: Percentage of first serves that are aces.
    \item \textbf{Serve points won $\%$}: Percentage of rallies won as server.
\end{itemize}

\begin{figure}[ht]
    \begin{center}
    \includegraphics[width=0.95\columnwidth]{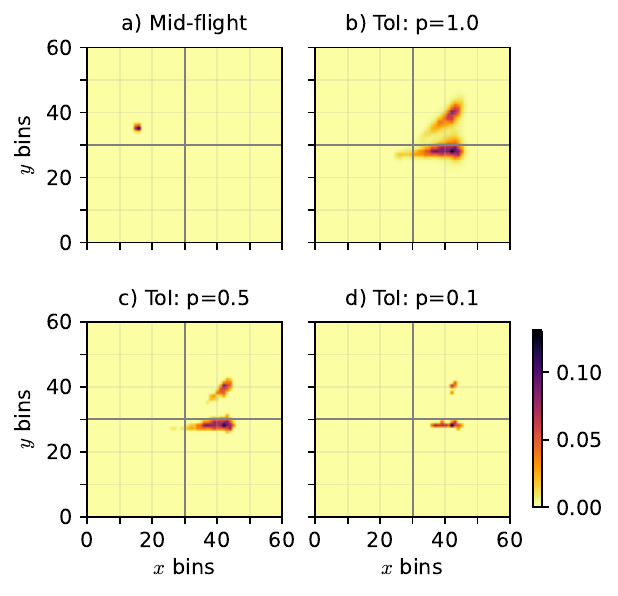}
    \caption{Bin probabilities for the ball projected into the $xy$ plane during a) mid-flight and b) - d) at time of impact (ToI) for 3 values of top-$p$.  The center of each diagram, bin $(x,y)=(30,30)$, corresponds to no movement. Yellow indicates a probability of 0 while progressively darker colors indicate higher probabilities.}
    \label{fig:output_probability_ball}
    \end{center}
\end{figure}

\paragraph{Varying the top-$p$ sampling parameter}
\cref{fig:output_probability_ball} shows typical output probability distributions, projected into the $xy$ plane, for an update step of the ball in mid-flight, and at the moment the ball is about to be hit.
The peaks in intensity for the mid-flight predictions (a) are distributed over very few bins since the model has learned the physical constraints of the system (e.g.\ drag, gravity), and can therefore be very confident in how to update the ball state.
The remaining panels (b)-(d) depict the probability distributions for the ball at the time of impact (ToI) \--- the point at which a player hits the ball for various values of top-$p$.
The distributions in these cases contain multiple separated peaks in intensity in the $xy$ plane.
This corresponds to different decisions a player may make when choosing to play the shot either down the line or across the court. By sampling from these two modes, we are able to perform counterfactual analysis (see \cref{sec:experiments}).
As top-$p$ decreases, the probability of sampling a cross court shot decreases, demonstrating the need to optimize top-$p$ in order to accurately capture the player behaviour in the simulations.
In general, we will see that a low top-$p$ value will result in less variety in playing style, but a high top-$p$ value will result in many outliers.

\begin{figure*}[ht]
    \centering
    \includegraphics[width=0.99\textwidth]{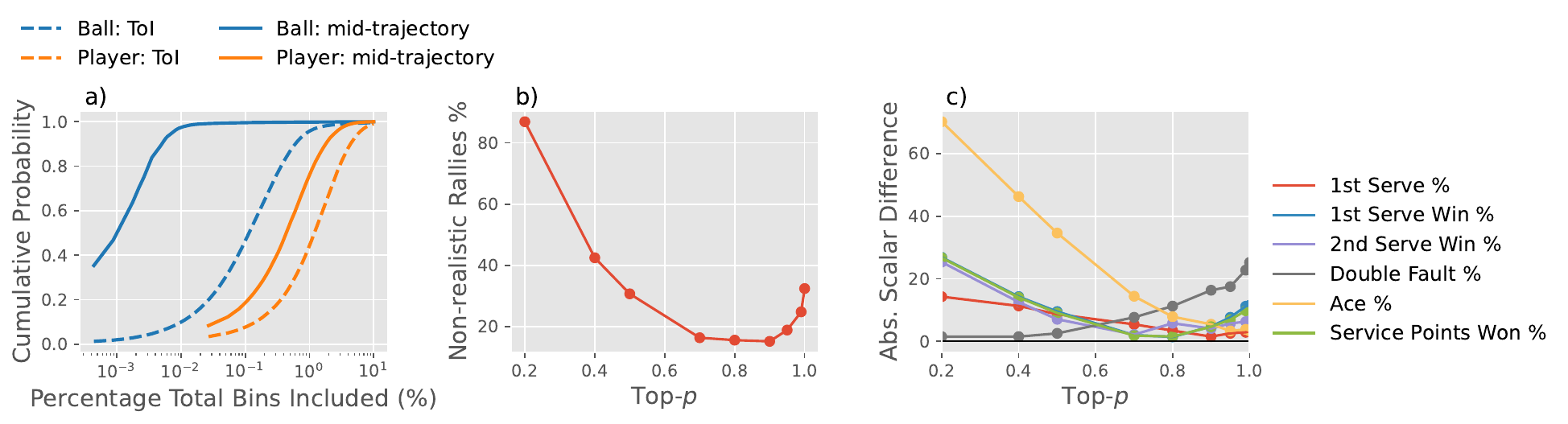}
    \caption{a) Cumulative probability for the player and ball at ToI for a return and during mid-flight for a shot as a percentage of number of contributing bins. b) Proportion of non-realistic rallies that are discarded during match simulation. c) Absolute difference between aggregated statistics in the training data and the simulations as a function of top-$p$.}
    \label{fig:sampling_parameter_tune_perc}
\end{figure*}

We see the similar patterns in \cref{fig:sampling_parameter_tune_perc}a), the cumulative probability for the player and ball at ToI for a return and during mid-flight for a shot as a percentage of number of contributing bins.
For mid-flight predictions of the ball, the probability distribution is concentrated over few bins, with 90$\%$ of the distribution contained within 0.002$\%$ of the total bins.
When predicting changes in direction (e.g. at ToI), the probability distribution is spread over more bins, up to 0.5$\%$ of the bins are required to populate 90$\%$ of the cumulative probability. 
\begin{figure*}[h!]
    \centering
    \includegraphics[width=0.8\textwidth]{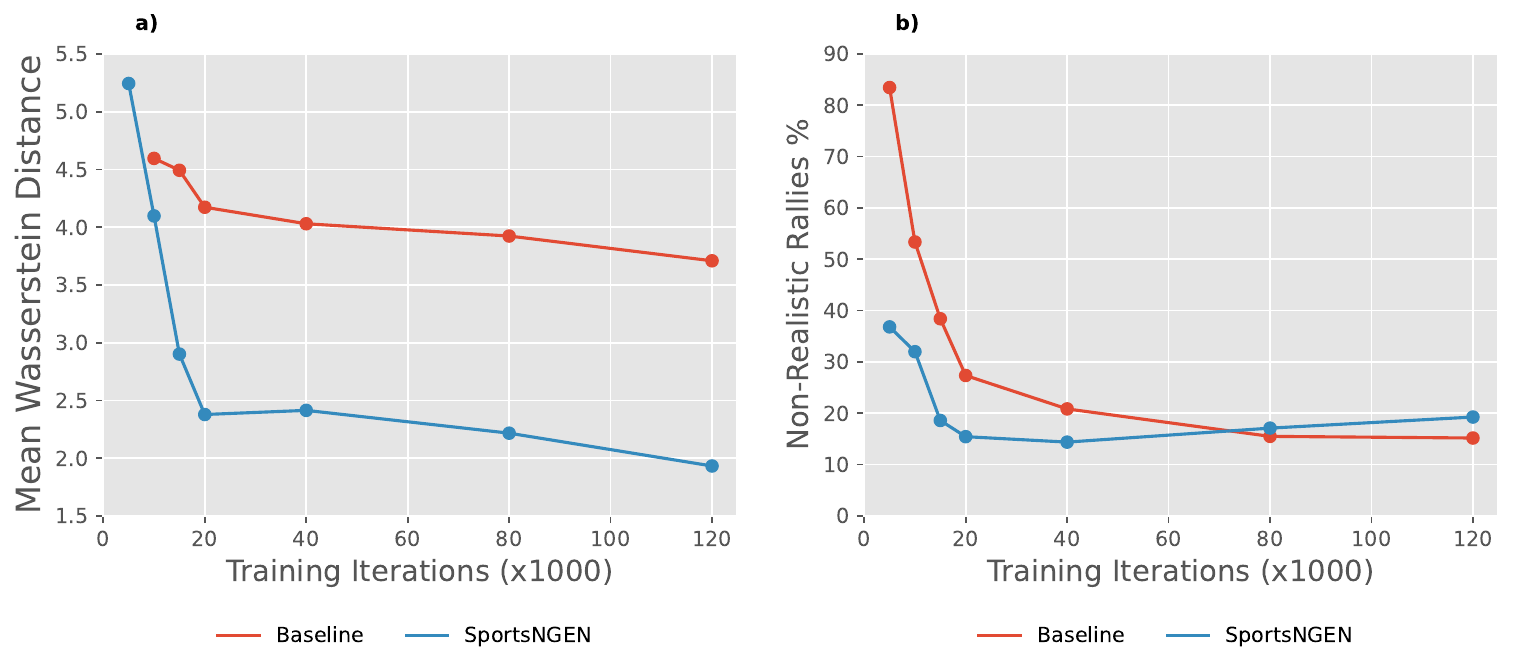}
    \caption{A comparison of convergence for \textit{SportsNGEN} against a Baseline model without $v$, $d$, $e$ and context tokens $C$, for a) An average of the 4 physical metrics, and b) Non-realistic rallies as a function of training iterations.
    }
    \label{fig:ablation_study}
\end{figure*}

\cref{fig:sampling_parameter_tune_perc}b) and (c) show how the various metrics vary with top-$p$. In (b), the number of non-realistic rallies (rallies that must be discarded based on logical checks) increases with a value of top-$p$ both that is too high, and too low.
Increasing top-$p$ increases the probability that the ball trajectory may be updated in a way that defies physical constraints and would be forced to be removed. With too low top-$p$ there may be too few options for the ball and player to update in a way that leads to a realistic rally.
In (c), we see that with the exception of double fault percentage, the metrics reach optimal values when top-$p$ is in the range of 0.8 to 0.9.
\paragraph{Object Token Component Ablation Study}
In \cref{fig:ablation_study}, we quantify how the additional components in the token vector $O$ affect the convergence and final accuracy of the physical metrics when compared to a baseline model that does not use velocity $v$, distance to the ball $d$, elapsed time $e$, or context tokens $C$ (similar to that used in \ballervec{}).
The plots show that \engine{} converges faster and reaches better results than the baseline model when averaged across all physical metrics.
We also see faster convergence to $\sim$20$\%$ non-realistic rallies.

\begin{figure}[h!]
    \centering
    \includegraphics[width=0.40\textwidth]{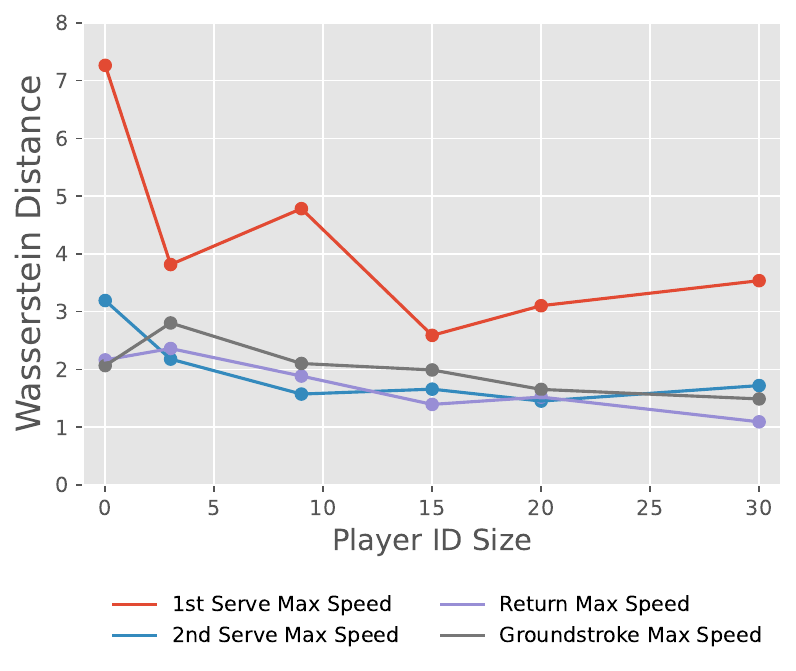}
    \caption{Varying the player ID $I$ size $\iota$ to show how various metrics can be improved with a larger $\iota$. As a control, we train a generic model without player ID $I$.
    }
    \label{fig:player_encoding}
\end{figure}

\begin{figure}[h]
    \centering
    \includegraphics[width=0.49\textwidth]{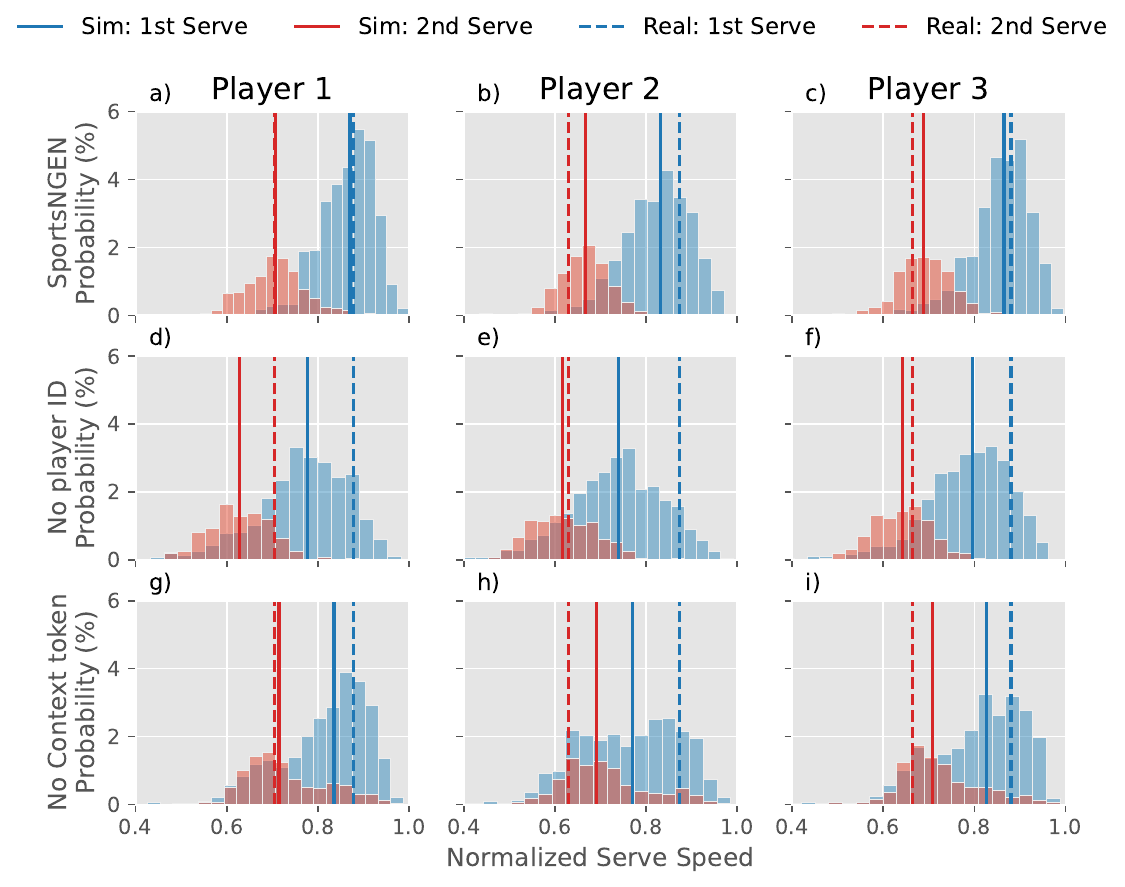}
    \caption{The distribution of first and second serve speeds for all three players for the following models: (top) \engine{}, (middle) a model with no player ID vector $I$, (bottom) a model with no serve context token $C_{serve}$.}
    \label{fig:serve_speeds_app}
\end{figure}

Varying the size $\iota$ of the player encoding vector $I$ in \cref{fig:player_encoding}, we find that the accuracy increases until $\iota=$20 where there are diminishing returns for further increases.
This is also supported by \cref{fig:serve_speeds_app} d)-f), where the data with no player ID $I$ has a much broader distribution of serve speeds, and a nearly identical median serve speed for all three players.

\paragraph{Context Token Study}
We add context tokens to encode the tournament, court surface type, and whether the serve is the player's first or second.
Typically the second serve is expected to be slower since players will prioritize accuracy over speed to avoid losing a point through double fault.
\cref{fig:serve_speeds_app} shows that the addition of a serve context token $C_{serve}$ as well as the player ID component $I$ in $O$ reduce the difference between real and simulated serve speeds and produce narrower distributions between first and second serve speeds.
\begin{figure}[h]
    \centering
    \includegraphics[width=0.49\textwidth]{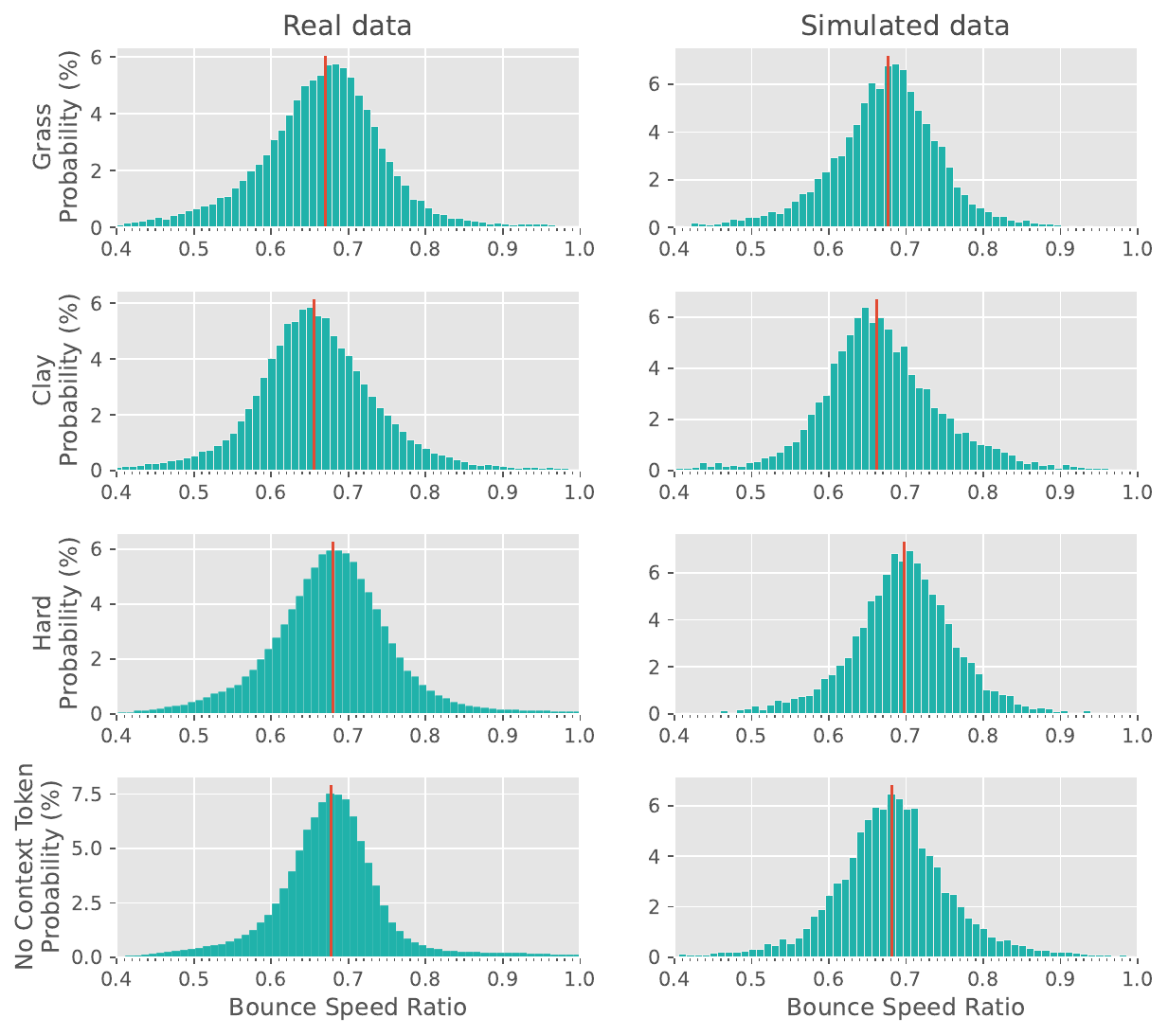}
    \caption{Ratio of speed after to before the ball bounce. Each row contains results for a different court surface type. The columns are real (left), and simulated (right) data. The last row is surface type agnostic, containing a weighted average of the data for each court.}
    \label{fig:bounce_plots}
\end{figure}

To quantify the effect of the playing surface, we use the coefficient of restitution by taking the ratio of the speed after to before the bounce.
A value less than 1 means the ball has lost momentum and indicates a slower surface.
\cref{fig:bounce_plots} shows this metric for three court types and for the surface agnostic case, for both real and simulated data.
The median value for each court type follows the expected trend: typically clay courts have the slowest bounces, and hard courts have the fastest, which is better represented when we introduce the surface token into the model.

\begin{figure}[h]
    \centering    
    \includegraphics[width=0.49\textwidth]{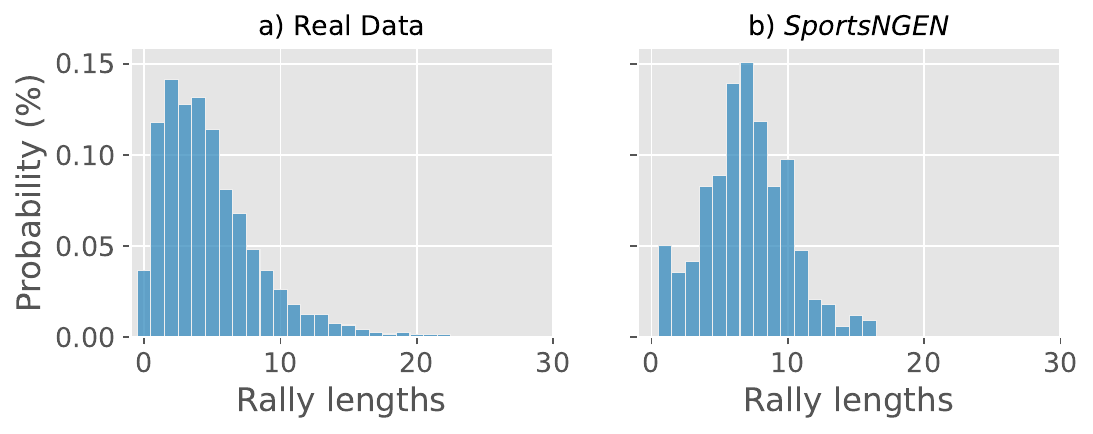}
    \caption{Length of rallies in number of shots for a) the original training data for the given three players on hard surfaces, b) simulated data using \textit{SportsNGEN}.}
    \label{fig:rally_lengths}
\end{figure}

We also demonstrate that \textit{SportsNGEN} is realistic throughout the rally with \cref{fig:rally_lengths} showing the distribution of rally lengths for real data, and simulations from \textit{SportsNGEN}.
Although we see a slightly higher peak in rally lengths in (b), we see both distributions with a peak at a small number of shots per rally, and tailing off towards 15 shots.

\paragraph{Transfer Learning}
As an extension to training a model capable of capturing the behavior of all players, we also train a generic model $f_{gen}$ which learns a single feature vector $I_{gen}$, called the \textit{generic player} vector where $I_n = I_{gen}, n \in N$.
We then fine-tune $f_{gen}$ with matches containing a specific player, and transfer learn a new set of $I_n \in N$ for the player that can represent their behavior against a generic opponent. This could be used for quickly customizing a pretrained model to a new player on the circuit.

\begin{figure}[h]
\begin{center}
\centerline{\includegraphics[width=0.49\textwidth]{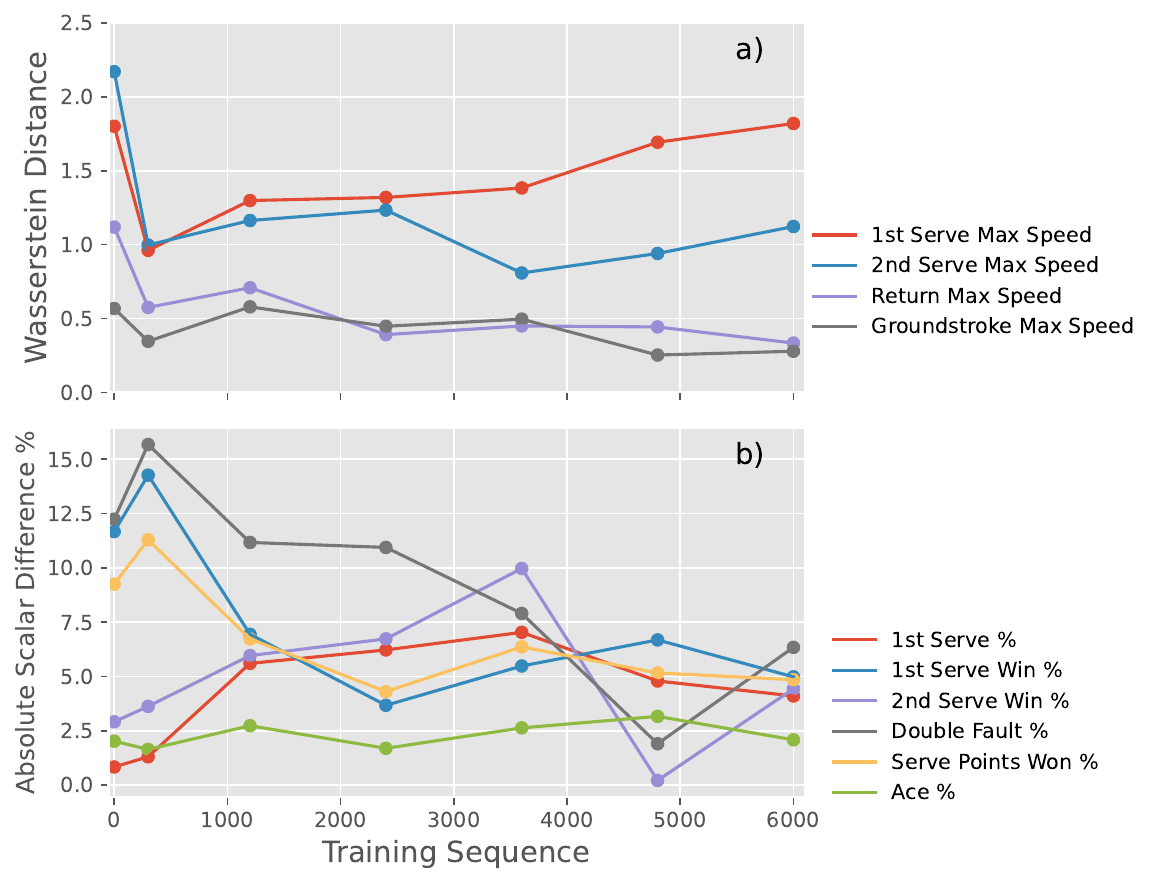}}
\caption{Learning features of a specific player by fine-tuning a generic model, showing a) the Wasserstein distance for physical data, and b) difference to training data for statistical metrics.}
\label{fig:transfer_learning_study}
\end{center}
\vskip -20pt
\end{figure}

\cref{fig:transfer_learning_study} shows various metrics as a function of the number of training sequences that are required to fine-tune $f_{gen}$ such that the generic player ID vector $I$ is adapted to a new player.
In the simulations, $f_{gen}$ is the opponent for the fine-tuned model.
The groundstroke and return metrics improve as the number of training samples increases whereas the serve metrics fluctuate with the first serve speed getting worse.
This can be explained by the low variability of the serve distribution being easier to learn when compared to highly variable groundstroke patterns.
\section{Applications}
In this section we explore \engine{} applications.

\paragraph{Predicting Rally Outcomes} 
A key intended application of \engine{} is generating insights for coaching and  sports broadcasts. To prove it's validity for these applications, the model should accurately forecast the probability that each player wins a rally as it develops.

We can test \engine{}'s ability to do this in the following way.
We sample random rallies from the training data, and roll out the model from a given random time step 100 times, to generate a win percentage for both players.
Repeating this for a large number of starting points, we form a histogram of predictions by stratifying the predictions into bins.

\begin{figure}[h!]
    \centering
    \includegraphics[width=0.35\textwidth]{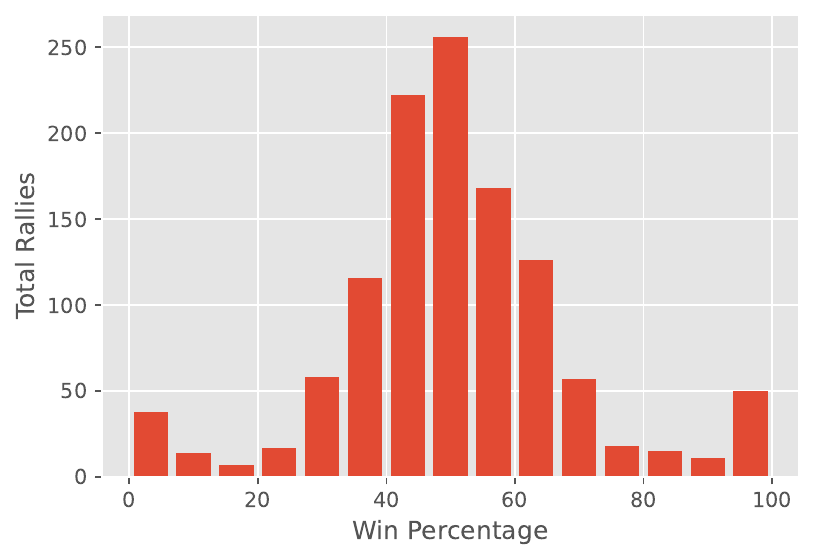}
    \caption{Histogram of win percentages output by the model when simulating rollouts in a random rally at a random point.}
    \label{fig:win_percentage_histogram}
\end{figure}

\cref{fig:win_percentage_histogram} shows the histogram of events contributing to the win percentage calibration plot. For each event, 100 simulated rollouts are used to generate the win percentage. The mean win percentage generated by the model is close to 50\% which is to be expected for tennis rallies. 
In addition there are situations in which the winner is very likely already determined (if the random time chosen is close to the end of the rally, for example). As a result, the bins close to 0 and 100\% are also more populated which explains the higher error in the more sparsely populated bins close to 20\% and 80\%.

\begin{figure}[h]
    \centering
    \includegraphics[width=0.40\textwidth]{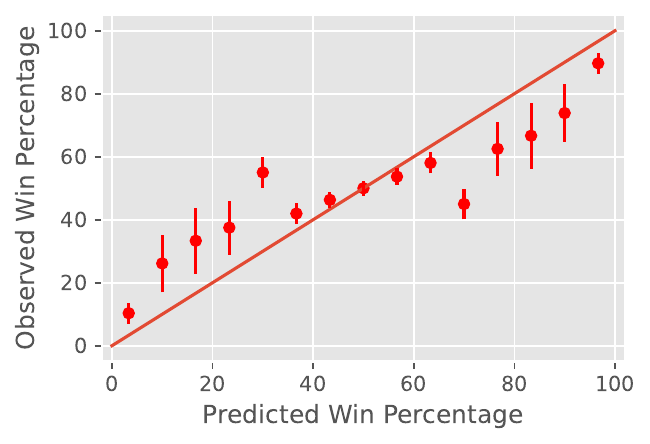}
    \caption{Predicted win percentages vs.~observed win percentages for \engine{}. The solid line shows ideal calibration. The win percentages output by \engine{} are well calibrated.}
    \label{fig:calibration}
\end{figure}

For each prediction, we also have the ground truth of who won the rally in the training data. Taking the 90\% bin for example, if the model is well-calibrated, the corresponding ground truth rallies should be won by the player in 90\% of cases.
\cref{fig:calibration} shows that the win percentages generated by the \engine{} are well-calibrated, with deviations where data are sparse. 
\paragraph{Counterfactuals}
\begin{figure*}[h!]
    \centering
    \includegraphics[width=0.80\textwidth]{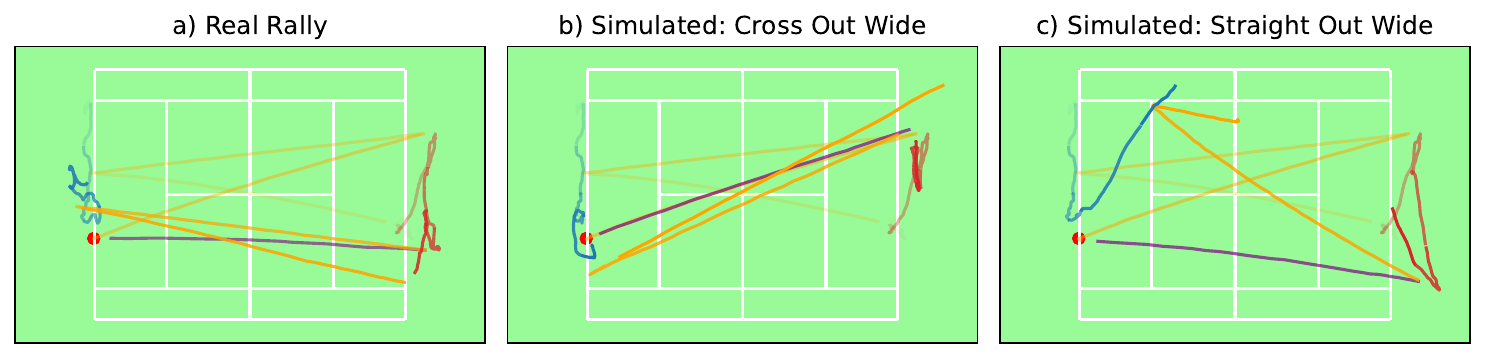}
    \caption{A real rally, and two simulated rallies for a different shot type, where the color transparency indicates time into the rally (with opaque being the end). The ball trajectory is orange, with the shot at which the simulations start shown in purple. The point at which the two simulations are branched is denoted by a red dot. The players are shown as blue and red traces.}
    \label{fig:rollouts_shot_decision}
\end{figure*}
\begin{figure*}[h!]
    \centering
    \includegraphics[width=0.9\textwidth]{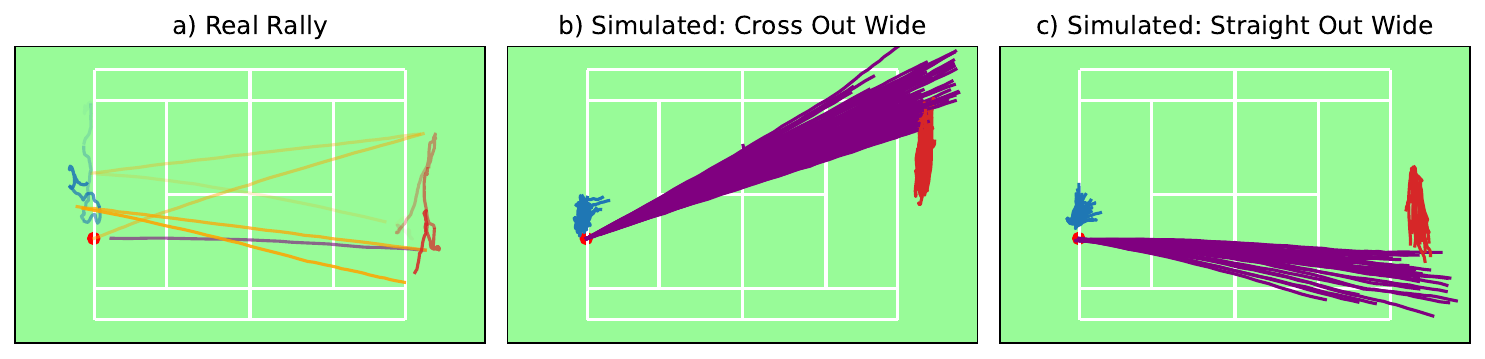}
    \caption{A real rally a), and many simulated rallies for two different shot types b), c). In the real rally, the increasing color opacity indicates time into the rally. The ball trajectory is orange, with the shot at which the simulations start shown in purple, the point at which this is branched is denoted by a red dot. The players are shown as blue and red traces. In the simulations, only the shots after the decision are shown to highlight the possibilities arising from the simulation engine.}
    \label{fig:rollouts_shot_decision_app}
\end{figure*}
\cref{fig:rollouts_shot_decision} demonstrates one way the \engine{} can be used to inform coaching decisions.
A point indicated by the red dot in a real rally is chosen as a branch point in time, just as a player is about to play a shot.
In the real rally, the shot after the branch point goes straight down the middle \--- indicated by the purple line in (a).

We can force alternative shot selection by sampling from the cross court mode in \cref{fig:output_probability_ball} and analysing how the rally would have played out. In (b) and (c), two alternatives are depicted. We quantify the strength of each shot selection by running 100 simulations until the end of the rally, sampling equally from both modes at this branch point. We then aggregate statistics to calculate a win percentage for each shot choice. 

Playing a shot to either of the two corners gave the player roughly equal probability of winning at 58$\%$, whereas the original choice of hitting to the middle reduced the probability below 50$\%$. Pushing the opponent farther to the edge of the court may explain this advantage.
\cref{fig:output_probability_ball} shows that the probability of the cross court mode is lower, highlighting that the player would more often opt for the safer shot down the middle.

\cref{fig:rollouts_shot_decision_app} shows the results of many simulations forcing a certain type of shot for the shot shown in purple. It shows that even if there are constraints imposed on the type of shot, there can still be variability in play.
Running this simulation for many shots and aggregating win percentages can give insight into the kinds of tactics that would be advantageous, and since the player and court can be specified and trained on real data, it could be specifically useful for improving the play style of a player in a particular situation. 

\paragraph{Football}
Though we focused this work on tennis, we have had success using \engine{} to simulate football matches with a high degree of realism using the same model architecture.
Click on \url{https://youtu.be/M0kkKiGVNzk} for a video demonstration of sustained passing sequences.
The player and ball positions are derived from COM data.
\section{Limitations}
An important limitation of SportsNGEN is that it is not designed to handle out of distribution situations.
Unconventional initial conditions can produce unreliable results.
This extends to unseen players in which the model will default to a “generic” player representation.
This method is also computationally intensive, requiring 2 days to train on an NVIDIA A100 GPU.
While we believe our method is applicable to many sports, we have only trained models for football and tennis.
Other sports may introduce difficulties.
\section{Discussion}
\label{sec:discussion}
In this work, we detailed \engine{} that is capable of generating realistic sports gameplay when trained on player and ball tracking sequences.
A unique aspect of the system is the ability to customize gameplay in the style of a particular player via fine-tuning.
Also, it is straightforward to use \engine{} to inform coaching decisions and game strategy through counterfactuals. 
In the future, we plan to adapt \engine{} to sports beyond tennis and football.
\section*{\uppercase{Acknowledgements}}
The authors would like to thank Beyond Sports B.V. for the visualisations and Sports Interactive for synthetic football data.
We also thank Anirban Mishra, Tristan Fabes, and Pavlo Sharhan for their helpful contributions.

\bibliographystyle{apalike}
{\small
\bibliography{references}}
\section*{\uppercase{Appendix}}
\begin{figure}[h]
    \centering
    \includegraphics[width=0.45\textwidth]{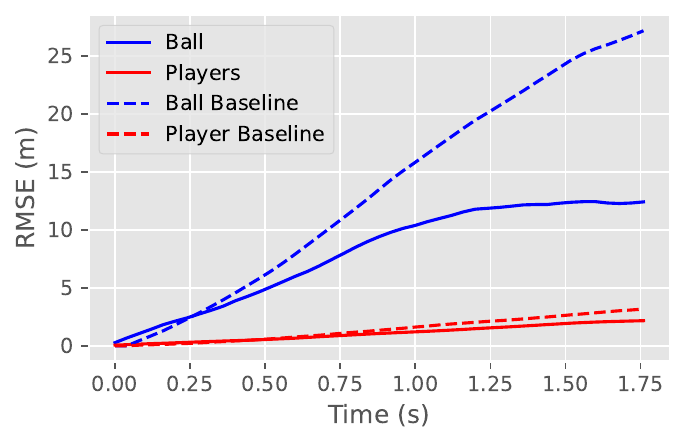}
    \caption{Root Mean Squared Error (RMSE) compared to real tennis data as a function of time, for both ball and player positions when simulating forward from a random in a rally. \engine{} performs better than a baseline of linear extrapolation.}
    \label{fig:mse_vs_time}
\end{figure}

\textbf{Prediction Error versus Time}
\cref{fig:mse_vs_time} shows the results from 200 simulations initialized from a random point in a random rally.
The simulations are evolved for 1.75 seconds and the RMSE is plotted compared with the ground truth data for the ball and players.
The baseline is taken as a linear extrapolation of the velocity of the player and ball frozen at the time the simulation begins.
Our simulation performs better than a linear extrapolation over a short time, indicating it has learned how to sensibly predict and update the state vectors as a function of time.

\end{document}